\documentclass[conference]{IEEEtran}
\usepackage{geometry}
\usepackage{graphicx}
\usepackage{times}
\usepackage{subcaption}
\usepackage{hyperref}
\usepackage{tcolorbox}
\usepackage{enumitem}

\begin{document}
%\IEEEoverridecommandlockouts

\title{\Large\bf An automated approach for improving the inference latency and energy efficiency
of pretrained CNNs by removing irrelevant pixels with focused convolutions}

\author{\normalsize
\begin{tabular}{ccccc}
Caleb Tung*                        & Nicholas Eliopoulos* & Purvish Jajal*    & Gowri Ramshankar*                         & Chen-Yung Yang*     \\
\small tung3@purdue.edu                   & \small neliopou@purdue.edu  & \small pjajal@purdue.edu & \small gramshan@purdue.edu                       & \small k8i5n6g20@gmail.com \\[5pt]
Nicholas Synovic\dag & Xuecen Zhang\textasciicircum        & Vipin Chaudhary\textasciicircum  & George K. Thiruvathukal\dag & Yung-Hsiang Lu*     \\
\small nsynovic@luc.edu                   & \small xxz1037@case.edu     & \small vipin@case.edu    & \small gkt@cs.luc.edu                            & \small yunglu@purdue.edu   \\[5pt]
\multicolumn{5}{c}{\small * Elmore Family School of Electrical and Computer Engineering, Purdue University, West Lafayette, Indiana, USA}              \\
\multicolumn{5}{c}{\small \dag Department of Computer Science, Loyola University Chicago, Chicago, Illinois, USA}                          \\
\multicolumn{5}{c}{\small \textasciicircum Computer and Data Sciences Department, Case Western Reserve University, Cleveland, Ohio, USA}                             
\end{tabular}
}

% 	\author{\normalsize
% 	\begin{tabular}[t]{c@{\extracolsep{8em}}c}
% 		\large Caleb Tung& \large Nicholas Eliopoulos & Purvish Jajal & Gowri Ramshankar & Chen-Yun Yang & Nicholas Synovic & Xuecen Zhang & Vipin Chaudhary & George K. Thiruvathukal & Yung-Hsiang Lu \\
% 		\\
% 		Purdue & Coauthor Department \\
% 		Purdue University e  & Coauthor Institute \\
% 		City, ST~~zipcode & City, ST~~zipcode\\
% 		Tel: 123-456-7890 & Tel: +81-3-333-1234\\
% 		Fax: 123-456-0987 & Fax: +81-3-333-5678\\
% 		e-mail: aaa@bbb.ccc.ddd & e-mail eee@ffff.ggg.hh\\
% \end{tabular}}

% use for special paper notices
%\IEEEspecialpapernotice{(Invited Paper)}

% make the title area
\maketitle

\makeatletter
\def\ps@IEEEtitlepagestyle{%
  \def\@oddfoot{\mycopyrightnotice}%
  \def\@evenfoot{}%
}
\makeatother
\def\mycopyrightnotice{%
  \begin{minipage}{\textwidth}
    \footnotesize
    ~ \hfill\\~\\
  \end{minipage}
  \gdef\mycopyrightnotice{}% just in case
}

{\small\bf Abstract---
Computer vision often uses highly accurate Convolutional Neural Networks (CNNs), but these deep learning models are associated with ever-increasing energy and computation requirements.
Producing more energy-efficient CNNs often requires model training which can be cost-prohibitive.
We propose a novel, automated method to make a pretrained CNN more energy-efficient without re-training.
Given a pretrained CNN, we insert a threshold layer that filters activations from the preceding layers to identify regions of the image that are irrelevant, i.e. can be ignored by the following layers while maintaining accuracy.
Our modified focused convolution operation saves inference latency (by up to 25\%) and energy costs (by up to 22\%) on various popular pretrained CNNs, with little to no loss in accuracy.
}

\section{Introduction}
\label{sec:intro}

\begin{figure*}
    \centering
    \begin{subfigure}[]{0.15\textwidth}
        \includegraphics[width=\textwidth]{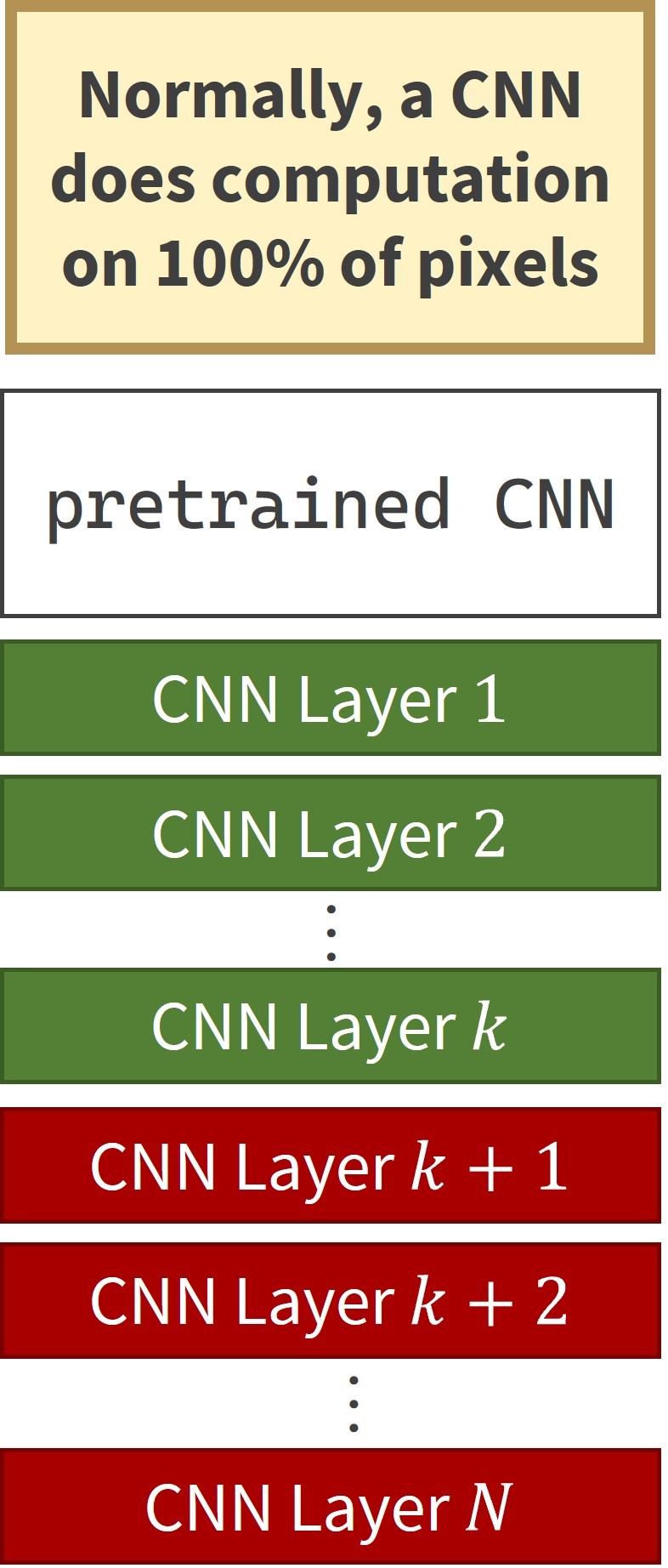}
        \caption{}
        \label{subfig:RegularCNN}
    \end{subfigure}     \hfill
    \begin{subfigure}[]{0.83\textwidth}
        \begin{subfigure}[b]{\textwidth}
            \includegraphics[width=\textwidth]{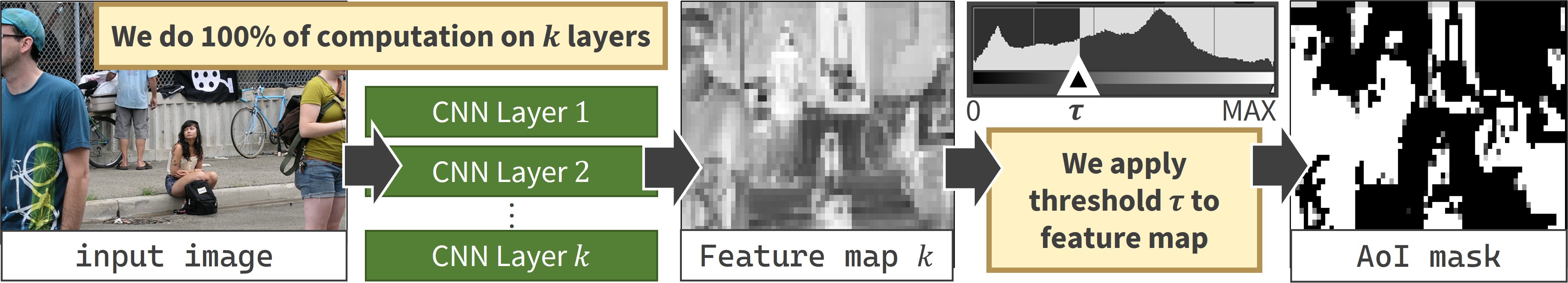}
            \caption{}
            \label{subfig:OurCNN1}
        \end{subfigure}
        \begin{subfigure}[b]{\textwidth}
        
            \includegraphics[width=\textwidth]{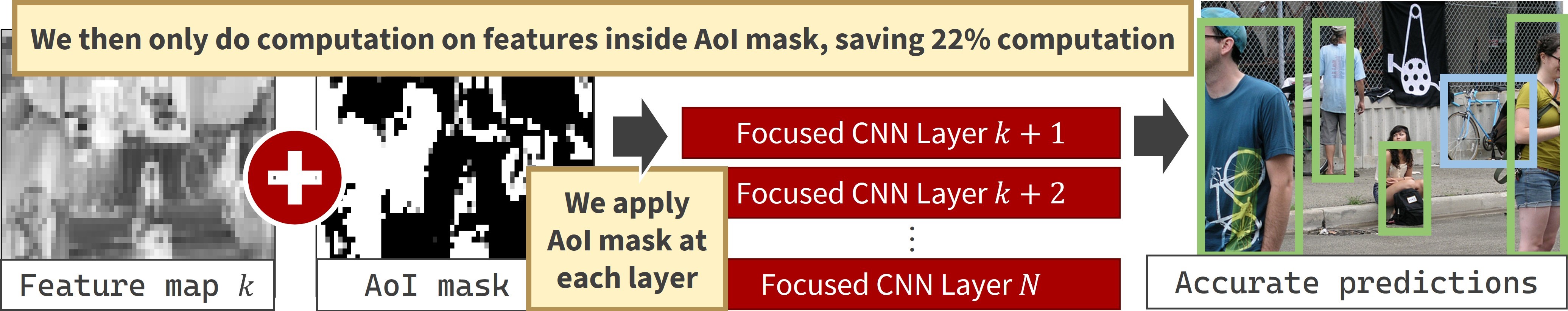}
            \caption{}
            \label{subfig:OurCNN2}
        \end{subfigure}
    \end{subfigure}
    \caption{
        \textbf{(a)} A pretrained CNN does computation on 100\% of the input pixels for all $N$ layers.
        The proposed method makes the CNN more efficient by:
        \textbf{(b)} First, only do 100\% computation during the first $k$ layers to collect contextual features.
        Second, apply a brightness threshold $\tau$ on the feature map from the $k$th layer to identify an Area of Interest (AoI) mask.
        As illustrated, white regions are relevant for an accurate prediction, black regions are irrelevant.
        Note: Select $k, \tau$ beforehand via a CNN energy consumption projection and an accuracy-vs-latency curve search, respectively.
        \textbf{(c)} Finally, in the last $N-k$ layers, completely ignore the irrelevant regions using focused convolutions.
        This saves computation with little to no loss in accuracy.
    }
    \label{fig:overview}
\end{figure*}

Pretrained Convolutional Neural Networks (CNNs) are prevalent because they enable anyone -- even those without the means to personally train CNNs -- to leverage state-of-the-art computer vision models.
CNNs are very computationally intensive and require power-hungry GPUs to execute, making them difficult to deploy in contexts like battery/solar-powered, mobile, embedded, and Internet-of-Things systems when GPUs or cloud offloading are unavailable.
Existing energy-efficient CNN techniques often require either an end-to-end retrain or a completely new model design.
Though successful, those techniques often require training and thus cannot be used with pretrained CNNs.

This paper proposes an easy-to-use method to \textit{reduce the energy consumption of pretrained CNNs without retraining.}
We insert a threshold layer into the pretrained CNN; this layer applies a brightness threshold to the activations from the preceding layers, determining which regions (background, uninteresting objects, etc.) of a given input image are deemed \textit{irrelevant} for an accurate inference.
The CNN's remaining layers are replaced with \textit{focused convolutions} which completely ignore those irrelevant regions, saving computation.
Illustrated in \autoref{fig:overview}, \textit{this method modifies a pretrained CNN using the following steps}:

First, to choose the $k$th layer at which to insert the threshold layer, we model the energy consumption of the layers in the CNN as a function of $k$, allowing us to automatically select the insertion point.

Next, to choose the brightness threshold $\tau$, we propose an automated latency-versus-accuracy curve search that yields a single threshold value to be used on the target dataset.  

Finally, we replace the remaining convolutional layers with our improved \textit{focused convolutions}~\cite{aicas} to ignore the irrelevant pixels.
The original focused convolution was a drop-in replacement for the standard General-Matrix Multiply (GEMM) convolutional technique used by contemporary AI libraries.
Made to only perform convolutions inside a predetermined Area of Interest (AoI) mask, their method required a different AoI mask per layer per inference along with a compute-heavy depth-mapping AoI generation technique.
This rendered their design less suitable for inference in environments where objects moved about or the scene changed regularly.
In contrast, our improvements use only one mask for the entire CNN and a hardware-aware block size, dramatically reducing computational overhead and boosting parallel processing utilization.

% Our novel design modifies a pre-trained CNN to generate Areas of Interest (AoIs) at \textit{inference time, using its existing training}, so that it can use focused convolutions on the AoI.
% No weights are changed; the improved CNN achieves either identical or mildly degraded accuracy relative to the original pretrained CNN while dramatically improving energy expenditure and inference speed.

\begin{tcolorbox}[colback=gray!10!white]
The proposed technique requires no training and instead modifies pretrained CNNs to save computation in resource-constrained deployments without GPUs or cloud offloading.

Our method automatically adjusts for different datasets and trades off accuracy, energy efficiency, and latency to meet a range of deployment requirements.

In some cases, considerable energy efficiency (up to 22\%) and latency improvements (up to 25\%) can be achieved without degrading accuracy.

A notable advantage of this training-free approach is that improvements can be achieved with little effort.
\end{tcolorbox}

We test the proposed technique on multiple popular pretrained models, including ResNet~\cite{resnet}, VGG~\cite{vgg}, ConvNeXt~\cite{convnext}, Faster-RCNN~\cite{fasterrcnn}, and SSDLite~\cite{ssdlite}.
We test on ImageNet~\cite{imagenet} for image classification and on Microsoft COCO ~\cite{coco} for object detection. Our method can reduce a pretrained CNN's inference energy consumption by up to $22\%$ on different processor types (Intel, AMD, Arm), with little to no loss of accuracy (0-2\% loss). Further, inference latency is shortened by up to $25\%$.

We also compare our technique's inference accuracy and latency with that of similarly inspired energy-efficient CNNs, \textit{showing that our method is either competitive or better than those techniques}, all without needing the training that the other techniques require.
 Code is open-sourced on GitHub at https://github.com/PurdueCAM2Project/focused-convolutions.

% In \autoref{sec:background}, we discuss related work on energy-efficient CNNs. In \autoref{subsec:irrelevant-pixels}, we provide background on what irrelevant pixels, Areas of Interest (AoIs), and focused convolutions are. In \autoref{sec:method}, we present the proposed technique and describe our design process. In \autoref{sec:results}, we test the proposed technique across varying pretrained CNNs, hardware configurations, datasets, and we discuss the results. The paper concludes in \autoref{sec:conclusion}.
% Code will be open-sourced on GitHub alongside publication.
\section{Related Work and Background}
\label{sec:background}
There are many efforts to improve CNN efficiency~\cite{goelsurvey}. They typically focus on: (1) reducing the model size (allowing it to fit more effectively into processor caches), (2) reducing the number of operations (reducing the load on the processor)~\cite{lpcvbook}.
Specialized hardware, such as neural accelerators~\cite{armethos}, can also run CNNs more efficiently. However, this paper focuses on software-side improvements.

\subsection{Similarly Inspired Methods to Our Technique}
\label{subsubsec:similarly-inspired}
Some techniques also use the concept of identifying unnecessary computation to skip at inference time. \textit{Early Exit architectures} decide if the CNN is confident enough to make a prediction before it finishes executing all the layers. The \textit{Spatially Adaptive Computation (SAC)} family of models can be trained end-to-end to identify regions of interest and then use special blocks to reduce computation outside those regions~\cite{sact}. We compare our technique with these methods in \autoref{sec:results}.

\subsection{Other Energy-Efficient, Low-Latency Methods}
\label{subsubsec:other-energy-efficient}
Traditional techniques make alterations to existing CNN architectures before training. \textit{Quantization} shrinks the size of a CNN by reducing the CNN's precision (e.g. storing numbers as 8-bit integers instead of 32-bit floats)~\cite{pytorchquant}. \textit{Pruning} reduces both computation and model size by deleting channels/features or neurons that are not activating often; the smaller model is then retrained~\cite{goelsurvey}. \textit{Knowledge Distillation} uses a large, trained model to ``teach'' a smaller,  more efficient model for deployment~\cite{goelsurvey}.
\textit{Neural Architecture Search} automatically searches a space of CNN building blocks for energy-efficient architecture.
% \textit{EfficientNet} specifically tries to minimize compute-intense blocks~\cite{efficientnet}. 
% \textit{Hierarchical Neural Networks} search for small neural networks configured in a tree, instead of a large, monolithic DNN.
% The root nodes eliminate parts of the search space, and the later neural network nodes in the tree refine the final prediction~\cite{hnn}.
% \textit{Restructurable Activation Networks} use a parameterized ReLU activation function to collapse entire sections of a CNN into a single lightweight, linear operation~\cite{ran}.

% \begin{figure}[t]
%     \centering
%     \includegraphics[width=\linewidth]{images/irrelevant-pixels.jpg}
%     \caption{Area of Interest (AoI) contains the Relevant pixels for a CNN to make a correct detection on the original image; other pixels are deemed Irrelevant and should be excluded from computation. As illustrated, Tung, et al.~\cite{aicas} generate AoIs using a depth-mapping neural network. This is computationally intensive.}
%     \label{fig:irrelevant-pixels}
% \end{figure}

\subsection{Irrelevant Pixels and Areas of Interest}
\textit{Irrelevant pixels} in an image do not contribute to the CNN's ability to make an accurate prediction.
Pixels that are useful to the CNN comprise \textit{``Areas of Interest'' (AoI)} of Relevant Pixels; irrelevant pixels fall outside the AoI~\cite{aicas}.
An image can have multiple, disjoint regions that comprise the image's AoI.

% \begin{figure}[t]
%     \centering
%     \includegraphics[width=\linewidth]{images/irrelevant-pixels3.jpg}
%     \caption{Existing method of using a depth-mapping neural network~\cite{aicas} (illustrated) to generate (Areas of Interest) AoIs, while accurate, are too computationally intensive. AoIs contain the Relevant pixels for a CNN to make a correct detection on the original image; other pixels are deemed Irrelevant and should be excluded from computation.  \textit{This paper inserts a layer into the CNN to filter out irrelevant pixels from computation.}}
  
%     \label{fig:irrelevant-pixels}
    
% \end{figure}

% Their method of generating AoIs uses a computationally intensive approach: they use a neural network called MiDAS~\cite{midas} to generate a depth map of an input image, and then filter the pixels of the depth map to find an AoI.  Their method is only suitable for stationary cameras (such as surveillance), it is impractical to generate AoIs on a per-image basis (e.g. for drone vision, detection datasets, etc.), because the overhead of AoI generation would outweigh the computational savings from ignoring the irrelevant pixels at inference time.
% Other AoI generation approaches include background subtraction~\cite{backgroundsub} and spectral residual saliency~\cite{saliency}.
The existing approach of generating AoIs using depth mapping neural networks~\cite{midas} is accurate but is too computationally expensive to deploy on resource-constrained systems.
% We first consider simple image processing approaches to identify AoIs before the CNN does inference. We conduct a brief study to explore these techniques. As shown in \autoref{fig:irrelevant-pixels}, AoIs can be generated using techniques like depth mapping~\cite{midas}, background subtraction~\cite{backgroundsub}, and spectral residual saliency~\cite{saliency}. We then test the generated AoIs with CNNs using focused convolutions to see if accuracy is negatively impacted. We find that depth mapping produces the best AoIs: CNNs achieve the highest accuracy using depth mapping. Unfortunately, the depth mapping approach is also compute-intensive since it requires the use of another neural network like MiDaS. Spectral residual saliency, while much faster, also produces the worst accuracy. Meanwhile, background subtraction techniques often produce AoIs that are too close to the image objects, obscuring contextual pixels needed to help the CNN make its predictions.
% Instead, this paper \textit{inserts a new layer that applies an activation brightness threshold $\tau$ to the feature maps generated by the CNN's early layers} (\autoref{sec:method}).

\subsection{How Focused Convolutions Work}
The \textit{focused convolution}~\cite{aicas} is based on the popular General-Matrix Multiply (GEMM) technique for doing convolutions~\cite{gemm}.
In GEMM, an input image is segmented into convolution kernel-sized \textit{patches}.
Each patch is then vectorized into a matrix's column or row. That matrix is then multiplied with the weights matrix to produce the convolution output.

% \begin{figure}[t]
%     \centering
%     \includegraphics[width=\linewidth]{images/gemm-conv.jpg}
%     \caption{In a standard GEMM convolution, the input image is converted into a matrix so that it can take advantage of parallelized hardware matrix multiplications to compute the convolution.}
%     \label{fig:gemm-conv}
% \end{figure}

The focused convolution, a drop-in-replacement for a GEMM convolution, applies an AoI that was generated in advance. Any patches not found inside the AoI are deemed irrelevant and then excluded from the matrix. That results in a smaller matrix and thus a less computation-intense matrix multiplication.
% \begin{figure}[t]
%     \centering
%     \includegraphics[width=\linewidth]{images/focused-conv.jpg}
%     \caption{Tung, et al's~\cite{aicas} focused convolution modifies the GEMM procedure as shown. When converting input patches (red, blue dotted line squares) into columns (red, blue dotted line columns) in the matrix, they filter out any patch that corresponds to irrelevant pixels in the AoI mask. This reduces the matrix size, allowing for a faster matrix multiplication operation.}
%     \label{fig:focused-conv}
The AoI not being generated at inference time keeps the CNN from being able to truly replace existing models for any application. 
% \textit{Our technique generates the AoI at inference time to solve this problem} (\autoref{sec:method}).

\subsection{Novelty of Our Contributions to Literature}
The related techniques described above all require varying degrees of training or otherwise completely redesigning the CNN. 
\textit{Our method, in contrast, requires no training and can be used on a pre-trained CNN.}
\section{Automatically Identifying and Removing Irrelevant Pixels at Inference Time}
\label{sec:method}
The proposed technique takes a pretrained CNN and then produces the \textit{fCNN}, a modified version of the CNN that applies a threshold to the early layers of the CNN to automatically generate AoIs and use focused convolutions for the remainder of the network with those AoIs. An fCNN behaves as follows:

\begin{enumerate}[noitemsep]
    \item Process input image using only the beginning $k$ layers ($k$ chosen in  \autoref{subsubsec:choosing_the_layers}) of the CNN (referred to as $NN.top$). Produces feature map $X$.
    \item Sum $X$ along the channels. Produces $X_{sum}$.
    % \nk{It might be better to stylize $NN.top$ in a different fashion - the spacing between N's looks odd with the math formatting. Or, we could use an underscore instead of a "$.$". Maybe \textit{NN.top} or $\textit{NN}_{\textit{top}}$}
    \item Filter $X_{sum}$ with the activation brightness threshold $\tau$ ($\tau$ chosen in \autoref{subsubsec:automated_dataset_aware}) to produce the AoI. Activations bright enough to clear the $\tau$ threshold are allowed through as corresponding regions of relevant AoI pixels; the rest are discarded. Produces $X_{thresh}$.
    \item All convolutional layers after $NN.top$ use our improved focused convolutions (\autoref{subsubsec:system_utilization_improvements}) on $X_{thresh}$, saving energy by discarding irrelevant pixels. The focused convolutions use the same weights and biases as the convolutions they replace.
\end{enumerate}

To generate this modified fCNN, we need to choose $k$ and $\tau$, and replace the convolutional layers after the $k$-th layer with focused convolutions.

% Our technique chooses $k$ by estimating the amount of energy consumed by the CNN and selecting the $k$  that allows the energy consumption estimate to satisfy a given target (e.g. 10\% energy savings). Next, the CNN is iterated a few times over the training dataset with different $\tau$ threshold values applied to the focused convolutions in the remaining layers, generating points on an accuracy-latency tradeoff curve for that CNN. The technique then chooses a $\tau$ that can satisfy the user's accuracy and latency targets $A$ and $T$, respectively. This newly configured fCNN can then be deployed for reduced energy consumption. While the concept bears similarities to attention mechanisms in some CNN and vision transformer models, the proposed method is more advantageous because it can modify a CNN without any training. The process is described in detail in the following subsections.

\subsection{Choosing the k Layers in NN.top}
\label{subsubsec:choosing_the_layers}
% \yhl{What does "must" mean here?}
The beginning $k$ layers of an fCNN will process every pixel of the input image, gathering information that can then be filtered by an inserted layer which applies activation brightness threshold $\tau$ to generate an AoI part-way through inference. This AoI will then be used by the later layers' focused convolutions.

The smaller $k$ is, the earlier the $\tau$-threshold will get applied. This also implies that more layers remain in the CNN to take advantage of the energy improvements of the focused convolution. Therefore, a smaller $k$ is beneficial.
Meanwhile, $k$ cannot be too small: too early on in the CNN,
there is insufficient information encoded within the features to make a useful threshold AoI~\cite{goelsurvey}. CNNs generally collect basic features about the input image in the first few layers. Deeper layers accumulate those into more complex features later in the network~\cite{vgg}. Therefore $k$ must be large enough to capture useful information in the features produced at the $k$th layer.

% \yhl{The following sentence is too vague and does not tell us anything useful. Please be more precise. What is "user"?}
To choose $k$, we use a heuristic: \textit{choose the latest layer (i.e., largest $k$) that is still small enough such that the resulting fCNN can meet the deployment energy requirement.}

% To choose one, we develop a heuristic that satisfies the following two requirements: (1) the layer must be \textit{late enough} in the CNN to have accumulated sufficient encoded information such that its output can indicate Areas of Interest, (2) the layer must still be \textit{early enough} in the CNN such that using its AoI in the following layers with focused convolutions can yield substantial compute savings.

% \textit{Ensuring the layer is late enough:}  Thus, we propose to use one of the layers in the vicinity of the first downsampling instance in the network (often, this is a maxpooling layer). This helps make sure the activations at that stage have encoded useful information for AoI generation.

Focused convolution energy savings were determined to be approximately linear with respect to AoI size~\cite{aicas} and CNN computation is known to scale linearly with respect to input size~\cite{yangetal}. Therefore, we linearly model the energy consumption of the CNN, and then use that information to project the energy savings of its fCNN equivalent, selecting $k$.

Let the energy consumption of the $i$th convolutional layer be $E_{c,i}$, the expected AoI size be $a$ as a percentage of the original input size, and the measured energy from the computational overhead introduced by the focused convolution be $c$ (this can be measured by manually setting the focused convolution AoI size to 100\% and then subtracting $E_{c,i}$). Then, the energy use of the corresponding focused convolutional layer $E_{f,i}$ is modeled linearly as $E_{f,i} = aE_{c,i} + c$. Thus, for a CNN with $k$ total conv layers, of which $k$ belong to $NN.top$, then there will be $N-k$ focused convolutional layers. We model the total energy consumed by the convolutions in the fCNN in \autoref{eqn:aoi_model}.

\begin{equation}
    E_{total} = (N-k)c + \sum_{i=1}^{k}E_{c,i} + \sum_{i=k+1}^{N}aE_{f,i}
    \label{eqn:aoi_model}
\end{equation}

Thus, $k$ can be selected such that $E_{total}$ just meets the deployment constraint. Note: if the deployment constraint is lower than the overhead costs, then our technique determines that a suitable fCNN is unachievable.

\subsection{Choosing Activation Brightness Threshold $\tau$}
\label{subsubsec:automated_dataset_aware}
The activation brightness threshold $\tau$ determines which pixels are considered relevant inside the AoI and which are deemed irrelevant. Our method tries the CNN with different $\tau$ values over a few iterations on the training dataset. After each full iteration over the dataset, it measures the average accuracy and inference latency to see if it meets the deployment requirements. If not, it iterates again with an adjusted $\tau$-value. Although this bears similarity to training, our method does not backpropagate or modify any model weights at all, whereas training requires many epochs and backpropagation~\cite{resnet}.

\begin{figure}[t]
    \centering
    \begin{subfigure}{\linewidth}
        \includegraphics[width=\linewidth]{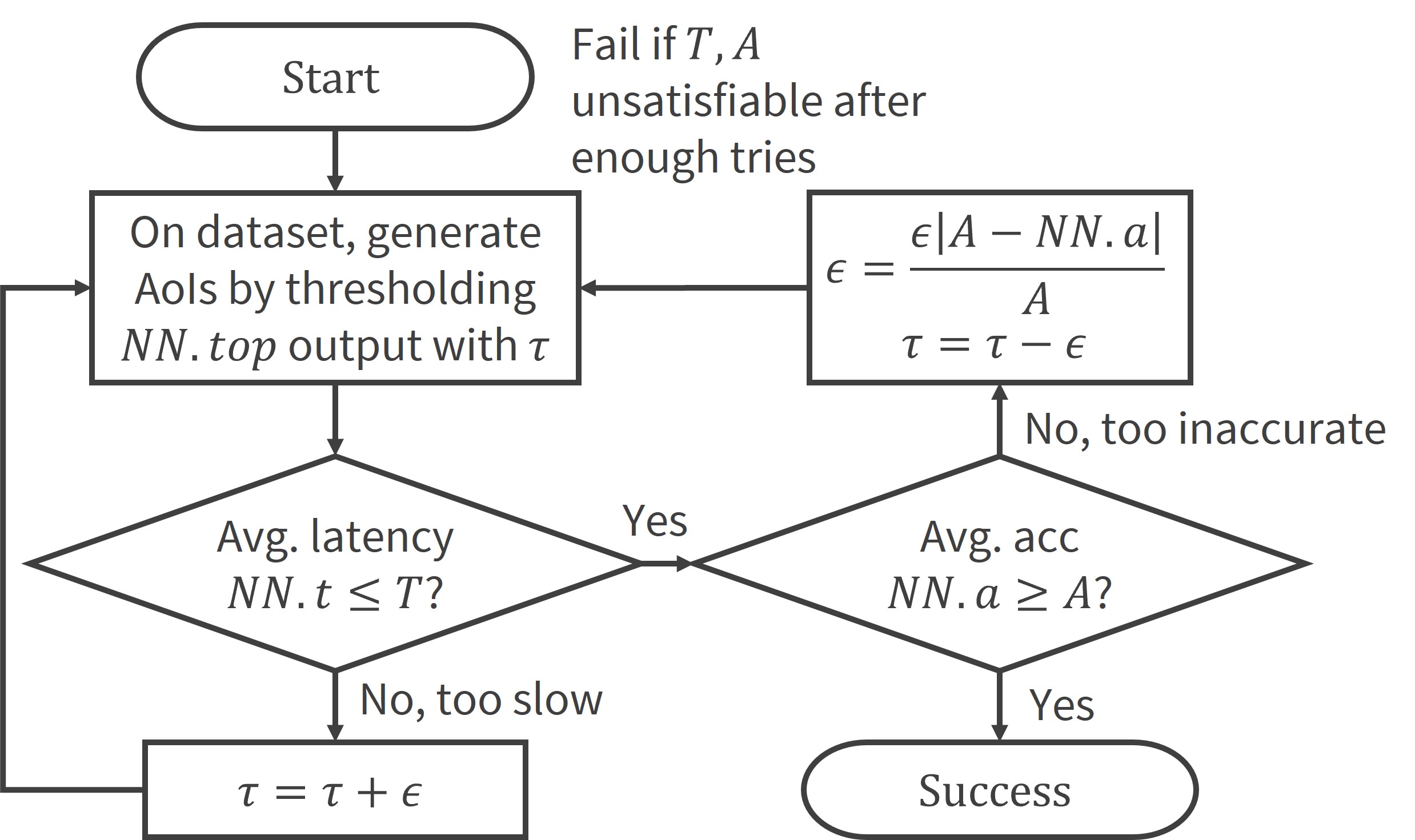}
        \caption{}
        \label{subfig:thresholdtuning}
    \end{subfigure}
    
    \begin{subfigure}{\linewidth}
        \includegraphics[width=\linewidth]{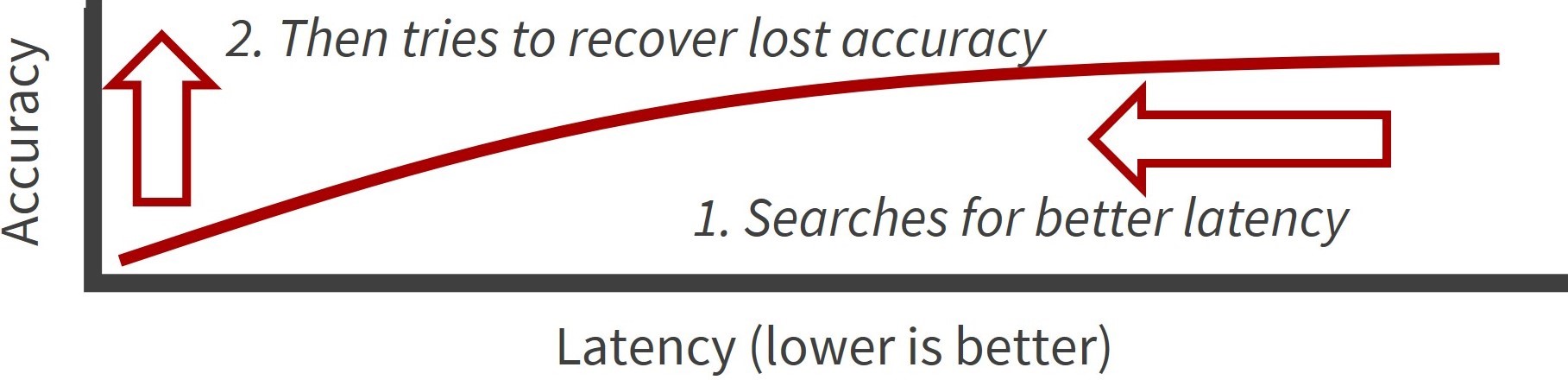}
        \caption{}
        \label{subfig:tradeoffcurvesearch}
    \end{subfigure}
    
    \caption{\textbf{(a)} Training-free process to choose activation brightness threshold $\tau$, given a maximum inference latency threshold of $T$ and a minimum accuracy threshold of $A$. This proposed method will succeed if latency and accuracy targets $T, A$ are simultaneously attainable, and fails otherwise. \textbf{(b)} This technique searches along the accuracy-latency curve, succeeding if $(T, A)$ is on the curve.}
    
    \label{fig:thresholdtuning}

\end{figure}

As shown in \autoref{fig:thresholdtuning}, the proposed technique chooses the activation brightness threshold $\tau$ as follows: $\tau$ is used to filter the output of $NN.top$, the top $k$ layers of the CNN. If a given region is brighter than the threshold, then it is allowed through as relevant pixels in the AoI. The higher $\tau$ is, the fewer pixels are allowed through and the smaller the AoI is (keep in mind that the AoI can be comprised of multiple distinct regions). $\tau$ is initialized to the minimum value of the sum of all $NN.top$ features, ensuring that any $NN.top$ output will pass through the threshold (i.e. 100\% AoI). We wish to achieve a maximum target $T$ for the CNN's inference latency $NN.t$, as well as a minimum target $A$ for the CNN's accuracy $NN.a$.

The proposed technique increases $\tau$ by increments of some $\epsilon$, shrinking the AoI and improving latency, until the latency target is met. Then, it checks to see if the accuracy target is also satisfied. If not, it begins reducing $\tau$ to attempt to find a smaller $\tau$ that can satisfy both targets. The size of the increment is adjusted based on the relative distance of $NN.a$ from $A$, getting smaller the closer the search gets to the target (i.e. as $|A-NN.a|$ shrinks in size, relative to $A$). The search succeeds if both $T, A$ are both attainable, and times out if the search cannot succeed after a pre-set period of time. Thus, the search explores along the accuracy-latency tradeoff curve, succeeding when $(T, A)$ is a point on or within the curve.

It is worth noting here that the $k$ selection process can be rolled into the $\tau$ selection framework: an outer loop can try different $k$ values, while the inner loop selects $\tau$ as shown. However, this paper recommends $k$ be pre-estimated using a simple mathematical calculation based on energy predictions because it achieves energy usage and latency improvements with fewer iterations over the training dataset. 

\subsection{Improving Focused Convolution Parallelization}
\label{subsubsec:system_utilization_improvements}

\begin{figure}[t]
  \centering
  \begin{subfigure}{0.3\linewidth}
    \includegraphics[width=\linewidth]{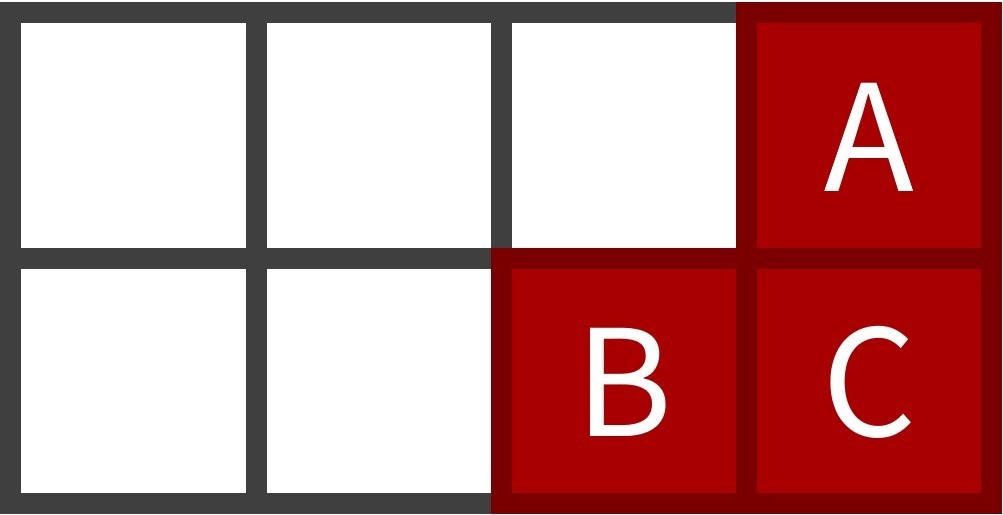}
    \caption{}
    \label{subfig:simd_1}
  \end{subfigure}
  \hfill
  \begin{subfigure}{0.3\linewidth}
    \includegraphics[width=\linewidth]{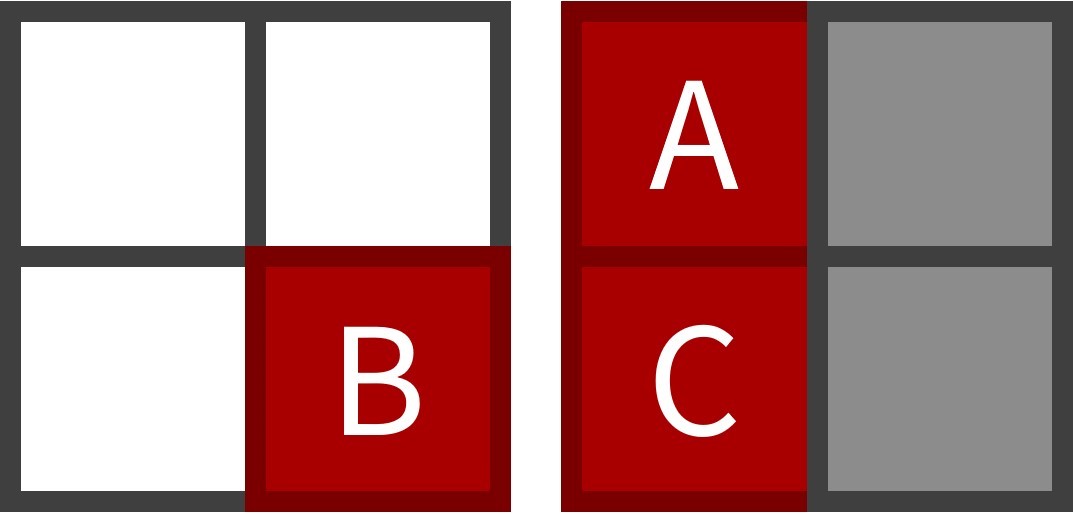}
    \caption{}
    \label{subfig:simd_2}
  \end{subfigure}
  \hfill
  \begin{subfigure}{0.3\linewidth}
    \includegraphics[width=\linewidth]{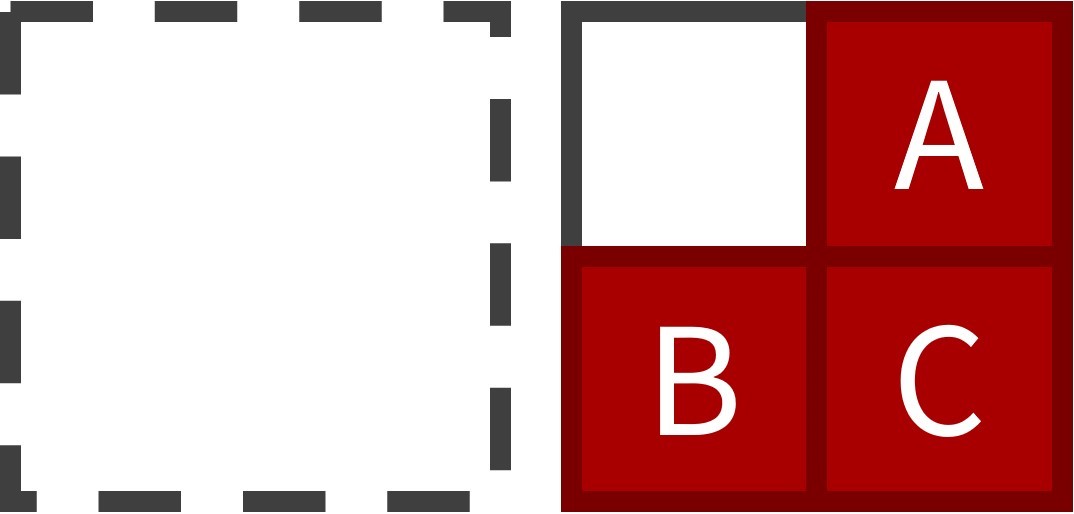}
    \caption{}
    \label{subfig:simd_3}
  \end{subfigure}
  \caption{Example of our technique's memory alignment. \textbf{(a)} $2\times4$ input data, where the three elements A,  B,C belong to the AoI. \textbf{(b)} On hardware that can parallelize the data processing in blocks of 4, the original focused convolution's sliding-window patch selection sends two blocks of size-4 data for processing: B is contained in the first block, while A, C are contained in the second block. \textbf{(c)} The proposed technique downsamples to multiples of the hardware blocksize, resulting in only one size-4 block of data being sent with A, B, C, thus saving processing on one block.
  }
  \label{fig:simd}
\end{figure}

% \begin{figure}[t]
%   \centering
%   \begin{subfigure}{0.3\linewidth}
%     \includegraphics[width=\linewidth]{images/frog.jpg}
%     \caption{}
%     \label{subfig:frog}
%   \end{subfigure}
%   \hfill
%   \begin{subfigure}{0.3\linewidth}
%     \includegraphics[width=\linewidth]{images/frog3x3.jpg}
%     \caption{}
%     \label{subfig:frog3x3}
%   \end{subfigure}
%   \hfill
%   \begin{subfigure}{0.3\linewidth}
%     \includegraphics[width=\linewidth]{images/frog16x16.jpg}
%     \caption{}
%     \label{subfig:frog16x16}
%   \end{subfigure}
%   \caption{Example of using a grid to choose hardware-friendly AoI. When the original image (\ref{subfig:frog}) produces an AoI as shown in (\ref{subfig:frog3x3}), the pixels are sent for processing based on a grid, where the cell size is selected in a hardware-friendly manner to match the block size of the device's parallel processors.}
%   \label{fig:frog_aoi_choosing}
% \end{figure}

Modern computer architectures implement specialized hardware (e.g., ``Neon'' vector registers on Arm CPUs and ``CUDA'' cores on NVIDIA GPUs) used to parallelize operations on multiple, independent pieces of data.
These parallel operations are supported by Single-Information-Multiple-Data (SIMD) instructions.
The data is collated into blocks that perfectly fit the size of a vector register on the processor.
The processor then uses specialized instructions to process the entire register’s data in parallel.
For example, a GEMM convolution operating on an entire input tensor could send multiple patches for simultaneous processing to improve inference latency.

Contemporary machine learning frameworks like PyTorch and TensorFlow already rely on C-native libraries to use parallel processing for built-in operations like ``Conv2d''.
The focused convolution already uses these libraries~\cite{aicas}.
However, we observe that the generic sliding-window indexing approach used by the original focused convolution results in inefficient utilization of the parallel processing cores. An example is illustrated in \autoref{fig:simd}: our technique will only select a single block of data instead of two, thus saving the processing.

To achieve this, we design the focused convolution to use a common computer architecture technique called \textit{memory alignment}~\cite{memoryalign}. We pre-divide the input into a grid, where each cell is sized in multiples of the parallel processing block size. If a cell contains part of the AoI, the entire cell is sent for parallel processing.
% An example is shown in \autoref{fig:frog_aoi_choosing}. 
Memory alignment ensures a more efficient usage of the processor than the original focused convolution. Because we do not need to directly change the kernel parallelization primitives, and only change the way the data is laid out, the proposed technique is instantly compatible with any libraries using optimizations for SIMD, CUDA, etc.

% \subsubsection{Additional focused convolution improvements}
% Previously, focused convolutions required a single forward pass of the CNN to pre-scale the AoI correctly, pixel-by-pixel, for the field of perception at each layer before the focused convolutions could be deployed. This is unacceptable overhead for AoI generation at runtime. By using memory alignment, our method can more efficiently compute which blocks to keep and which to ignore from a single AoI, removing the overhead cost of pre-scaling.
\section{Results and Discussion}
\label{sec:results}
% \begin{figure*}[t]
%     \centering
%     \includegraphics[width=\linewidth]{images/imgclsaccuracy.jpg}
%     \caption{For a pretrained CNN equipped with focused convolutions, as we vary the activation brightness threshold $\tau$, different amounts of activations are filtered, causing the CNN's accuracy, latency, and AoI size to change. We designate the CNNs that achieve the best accuracy as ``fCNNs''. Left: ImageNet accuracy trades off with inference latency. Fortunately, the tradeoff can be little to nonexistent -- our fVGG-16 achieves faster inference than VGG-16 without losing accuracy. Right: Larger AoI sizes yield better accuracy. Note that latency is reported from an AMD desktop CPU.
%     % Left: ImageNet Accuracy vs AMD CPU Latency, explored by the automated technique to choose the activation brightness threshold $\tau$. The focused-convolution models (denoted with ``f'') and the unmodified models are labeled for comparison. As shown, fCNNs can be selected such that they \textbf{match the accuracy of the original models, with better inference latency}. Right: Accuracy vs. AoI Size, showing that maximum ImageNet accuracy is achieved when the AoI is generated between 85\% and 90\% the image size, on average.
%     % % Accuracy appears to degrade linearly as the AoI shrinks, while inference latency improves.
%     % \todo[inline]{YHL: This figure looks nicer if you move the labels a little away from the data. Right now, they are too close.}
%     }
%     \label{fig:imgclsaccuracy}
% \end{figure*}

\begin{table*}[t]
\centering

\begin{tabular}{|l|l|r|r|rrr|rrr|}
\hline
\textbf{CNN}           & \textbf{Dataset}     & \multicolumn{1}{l|}{\textbf{Accuracy}} & \multicolumn{1}{l|}{\textbf{MAC/inf}} & \multicolumn{3}{l|}{\textbf{Energy/inference (J)}}                                                                 & \multicolumn{3}{l|}{\textbf{Latency/inference (ms)}}                                                                 \\ \hline
                       &                      & \multicolumn{1}{l|}{}                  & \multicolumn{1}{l|}{}                 & \multicolumn{1}{l|}{\textbf{Intel}} & \multicolumn{1}{l|}{\textbf{AMD}}  & \multicolumn{1}{l|}{\textbf{Arm}} & \multicolumn{1}{l|}{\textbf{Intel}}  & \multicolumn{1}{l|}{\textbf{AMD}}   & \multicolumn{1}{l|}{\textbf{Arm}} \\ \hline
VGG-16                 & ImageNet-1K          & 0.716                                  & 15.50G                                & \multicolumn{1}{r|}{6.9}            & \multicolumn{1}{r|}{11.2}          & 10.1                              & \multicolumn{1}{r|}{242.1}           & \multicolumn{1}{r|}{83.2}           & 2020.9                            \\
\textbf{fVGG-16}      & \textbf{ImageNet-1K} & \textbf{0.716}                         & \textbf{14.19G}                       & \multicolumn{1}{r|}{\textbf{6.4}}   & \multicolumn{1}{r|}{\textbf{10.9}} & \textbf{8.9}                      & \multicolumn{1}{r|}{\textbf{222.8}}  & \multicolumn{1}{r|}{\textbf{77.1}}  & \textbf{1799.0}                   \\ \hline
ResNet-18              & ImageNet-1K          & 0.698                                  & 1.82G                                 & \multicolumn{1}{r|}{2.0}            & \multicolumn{1}{r|}{3.1}           & 2.3                               & \multicolumn{1}{r|}{54.2}            & \multicolumn{1}{r|}{18.2}           & 457.7                             \\
\textbf{fResNet-18}   & \textbf{ImageNet-1K} & \textbf{0.697}                         & \textbf{1.60G}                        & \multicolumn{1}{r|}{\textbf{1.5}}   & \multicolumn{1}{r|}{\textbf{2.6}}  & \textbf{2.1}                      & \multicolumn{1}{r|}{\textbf{50.19}}  & \multicolumn{1}{r|}{\textbf{16.4}}  & \textbf{410.8}                    \\ \hline
ConvNeXt-T             & ImageNet-1K          & 0.821                                  & 4.47G                                 & \multicolumn{1}{r|}{3.4}            & \multicolumn{1}{r|}{6.5}           & 5.1                               & \multicolumn{1}{r|}{112.4}           & \multicolumn{1}{r|}{41.6}           & 960.4                             \\
\textbf{fConvNeXt-T}  & \textbf{ImageNet-1K} & \textbf{0.818}                         & \textbf{4.05G}                        & \multicolumn{1}{r|}{\textbf{2.9}}   & \multicolumn{1}{r|}{\textbf{5.2}}  & \textbf{4.3}                      & \multicolumn{1}{r|}{\textbf{99.9}}   & \multicolumn{1}{r|}{\textbf{37.0}}  & \textbf{854.3}                    \\ \hline
Faster-RCNN            & COCO                 & 0.370                                  & 120.87G                               & \multicolumn{1}{r|}{68.1}           & \multicolumn{1}{r|}{106.8}         & -                                 & \multicolumn{1}{r|}{2390.1}          & \multicolumn{1}{r|}{751.9}          & -                                 \\
\textbf{fFaster-RCNN} & \textbf{COCO}        & \textbf{0.370}                         & \textbf{101.30G}                      & \multicolumn{1}{r|}{\textbf{57.7}}  & \multicolumn{1}{r|}{\textbf{88.9}} & \textbf{-}                        & \multicolumn{1}{r|}{\textbf{2011.5}} & \multicolumn{1}{r|}{\textbf{616.6}} & \textbf{-}                        \\ \hline
SSDLite                & COCO                 & 0.210                                  & 716.42M                               & \multicolumn{1}{r|}{3.0}            & \multicolumn{1}{r|}{7.2}           & 5.7                               & \multicolumn{1}{r|}{100.5}           & \multicolumn{1}{r|}{48.6}           & 1083.7                            \\
\textbf{fSSDLite}     & \textbf{COCO}        & \textbf{0.192}                         & \textbf{599.06M}                      & \multicolumn{1}{r|}{\textbf{2.3}}   & \multicolumn{1}{r|}{\textbf{5.8}}  & \textbf{4.5}                      & \multicolumn{1}{r|}{\textbf{79.4}}   & \multicolumn{1}{r|}{\textbf{39.7}}  & \textbf{876.4}                    \\ \hline
\end{tabular}
\caption{Pretrained CNNs are compared with their corresponding ``fCNNs'' \textbf{(in bold)} using our method on an Intel laptop, an AMD desktop, and an Arm embedded device. Latency improvements can be achieved with little to no accuracy loss. Cells with ``-'' indicate that the model could not run on the device (exceeded memory capacity).}
\label{tab:results}
\end{table*}

% \ns{Can this NOTE be a footnote, or relocated after the conclusion?}

To demonstrate the utility of our method, we measure various performance aspects of different pretrained models when modified using our technique. We choose three pretrained image classifier models (VGG-16, ResNet-18, ConvNeXt-T) and two pretrained object detection models (Faster-RCNN and SSDLite) from Meta AI Research's Torchvision library. We then use the proposed technique, with ImageNet for image classification and Microsoft COCO for object detection, to determine $k$ and $\tau$. Then, we modify each model to use focused convolutions, and we measure energy consumption, inference latency,  accuracy, and Multiply-Accumulate (MAC), comparing the focused convolution models with the unmodified ones.

Note: In this section, we often compare the perfomance of an unmodified, pretrained CNN with the focused-convolution fCNN version using ``\% improvement'' or ``\% degradation''. This is calculated using the formula $|unmodified-focusedconv|/unmodified$.

\subsection{Experimental Setup}
We test using three devices with different levels of power consumption and different operating systems. We demonstrate improvements in energy consumption and inference latency without using any GPUs, hardware accelerators, or cloud offloading:

\begin{itemize}
    \item Embedded (5 W): Broadcom Arm Cortex-A5, Debian
    \item Laptop PC (28 W): Intel Core i7, Ubuntu
    \item Desktop PC (142 W): AMD Ryzen 9, Windows
\end{itemize}

On the Arm embedded device, energy consumption is physically measured using a Monsoon Solutions HV Power Monitor. On the Intel laptop PC, measurements are taken using Intel's ``Power Gadget'' software, and on the AMD desktop, measurements are recorded with ASUS' ``Armoury Crate'' software. Baseline steady-state power consumption is recorded and subtracted from the numbers measured during inference.

The focused convolutions are compiled to use with Pytorch. On each device, the SIMD blocksize \autoref{sec:method} is set according to the specifications from the processor's documentation. 

We measure inference accuracy and MAC on ImageNet and Microsoft COCO using averaged numbers from the ``torchbench'' and ``ptflops'' libraries.

We report accuracy using the Top-1 ImageNet accuracy metric~\cite{imagenet} and the Box mAP COCO accuracy metric~\cite{coco}.
% \ns{I disagree with using the Python timing, as there is significant overhead with using that library. Unix provides a "time" command and Windows has a similar feature exposed through powershell called "Measure-Command". Python's "timeit" module could also useful here} 

\subsection{Activation Brightness Threshold Selection}

We follow the method described in \autoref{sec:method} to create fCNNs from the pretrained CNNs and select $k$ and $\tau$.

For each model, several convolutional layers come before the first downsampling point, at which the input size shrinks due to either striding or pooling.
$k$ is selected at the first downsample point of the model.
This allows for enough flexibility to choose a $\tau$ while retaining sufficient image-wide information from the early layers. 

For image classification, we start the automated $\tau$ search for a latency target $T$ that is 10\% better than the pretrained CNN, with an accuracy target $A$ matching the accuracy of the original CNN.
As the $\tau$ value increases, less pixels are allowed past the threshold, shrinking the size of the AoI.
This also causes the model's accuracy to begin dropping linearly. However, there are cases (e.g. VGG-16) where the accuracy holds steady while latency drops, indicating that \textbf{it is possible to achieve the accuracy of the original models while improving latency.}

The technique selects the best point, where the most latency is saved while dropping the least accuracy.
Those models using our technique are denoted as the ``fCNN'' models.
The same process is repeated to determine the ``fCNN'' models for the object detectors on Microsoft COCO.

To choose the $\tau$ for our ``fCNN'' models, we do not need to retrain. Although our curve search will iterate over the training dataset, it is much faster than retraining a model, since we do not do backpropagation and only use 7 iterations to select a $\tau$ for each model.

\subsection{Improvements On Pretrained CNNs}
We compare our ``fCNN'' models with their corresponding unmodified pretrained CNNs in \autoref{tab:results}. As shown, across desktop, laptop, and embedded processors, the technique successfully converts pretrained CNNs into faster, more energy-efficient models that still achieve the same or mildly degraded accuracy. Object detection models achieve more improvements because the COCO images often have smaller AoIs than the ImageNet images.

\begin{figure}[t]
  \centering
      \begin{subfigure}{0.32\linewidth}
        \includegraphics[width=\linewidth]{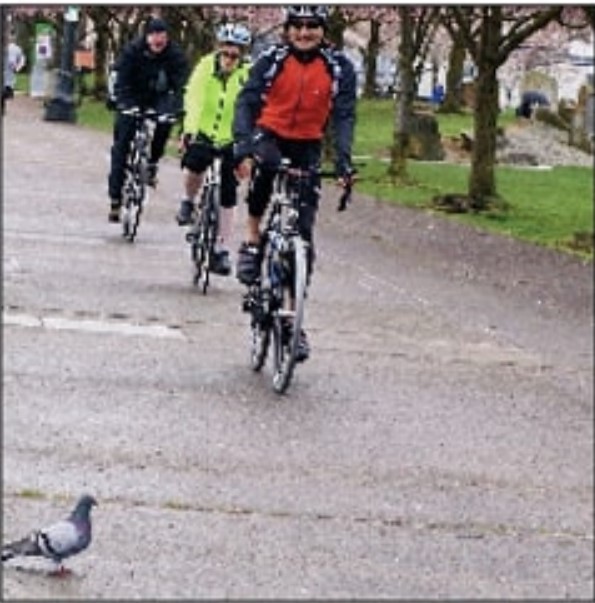}
        \caption{}
        \label{subfig:coco-base}
      \end{subfigure}
      \hfill
      \begin{subfigure}{0.32\linewidth}
        \includegraphics[width=\linewidth]{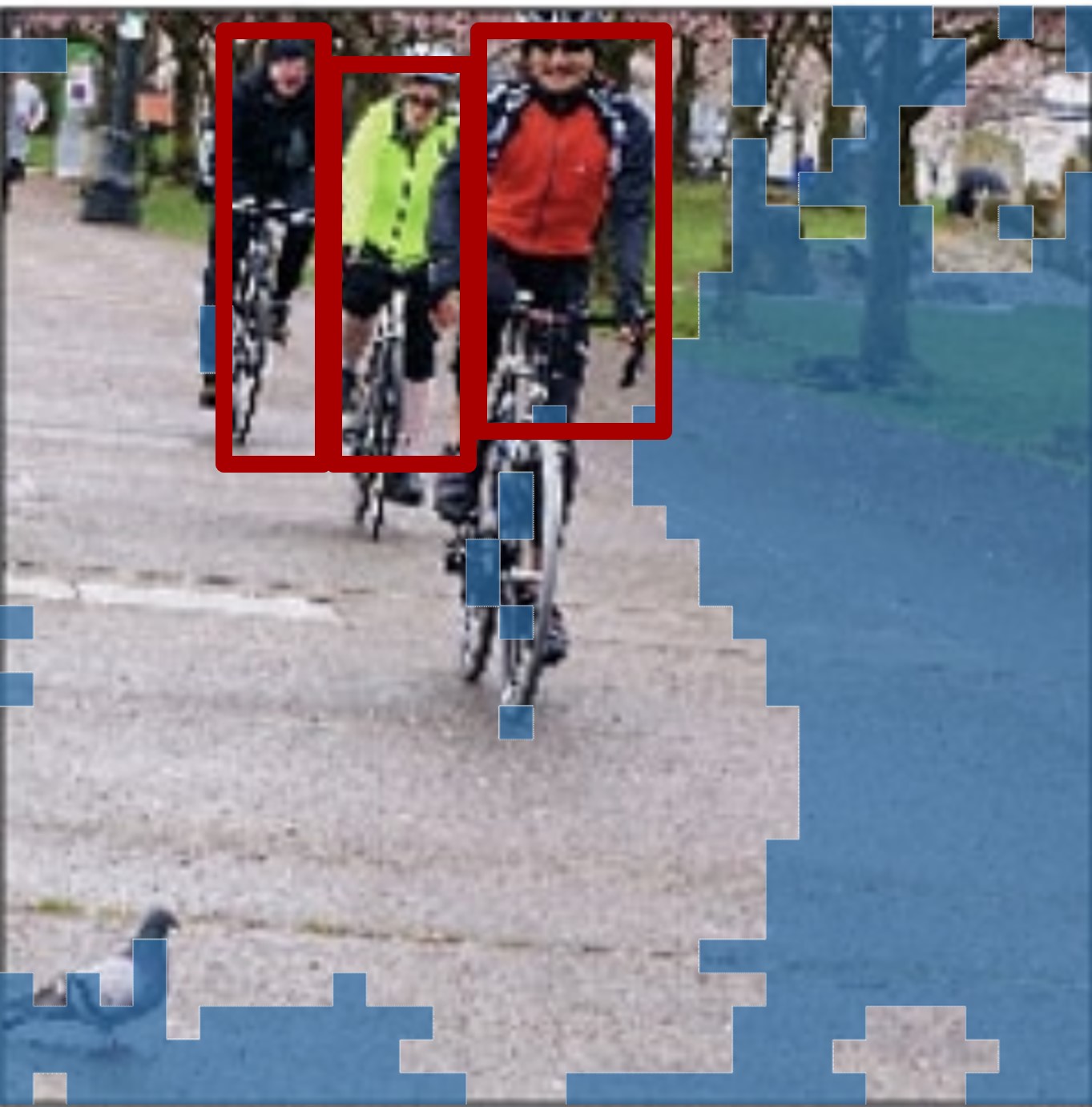}
        \caption{}
        \label{subfig:coco-fasterrcnn}
      \end{subfigure}
      \hfill
      \begin{subfigure}{0.32\linewidth}
        \includegraphics[width=\linewidth]{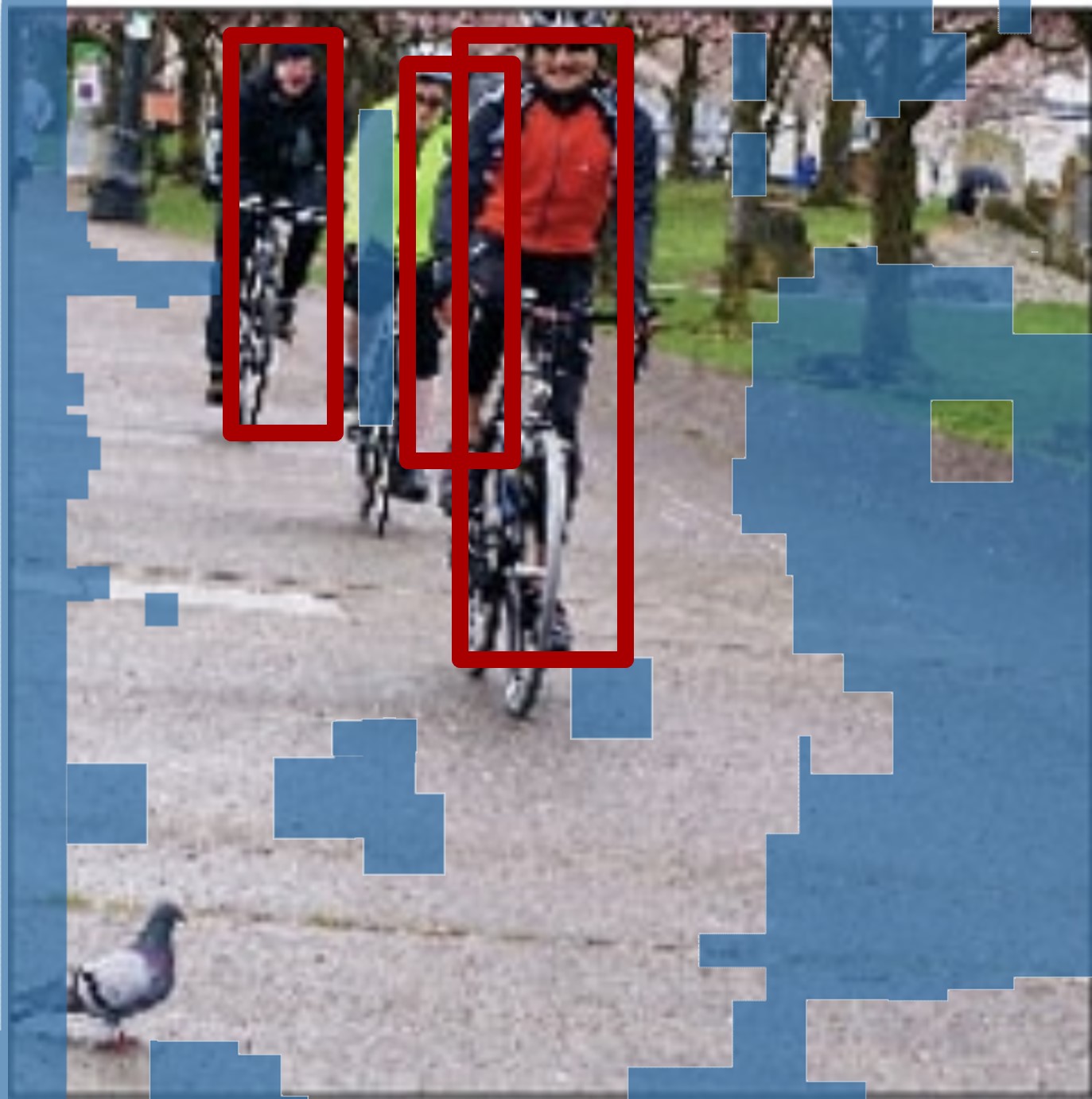}
        \caption{}
        \label{subfig:coco-ssdlite}
      \end{subfigure}
      
      \begin{subfigure}{0.32\linewidth}
        \includegraphics[width=\linewidth]{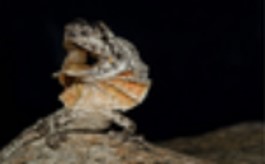}
        \caption{}
        \label{subfig:imagenet-base}
      \end{subfigure}
      \hfill
      \begin{subfigure}{0.32\linewidth}
        \includegraphics[width=\linewidth]{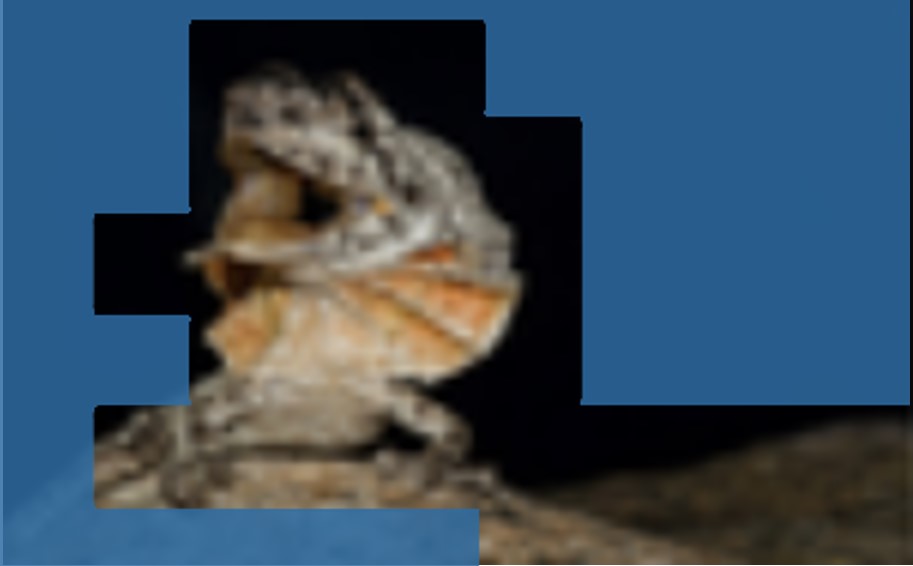}
        \caption{}
        \label{subfig:imagenet-resnet}
      \end{subfigure}
      \hfill
      \begin{subfigure}{0.32\linewidth}
        \includegraphics[width=\linewidth]{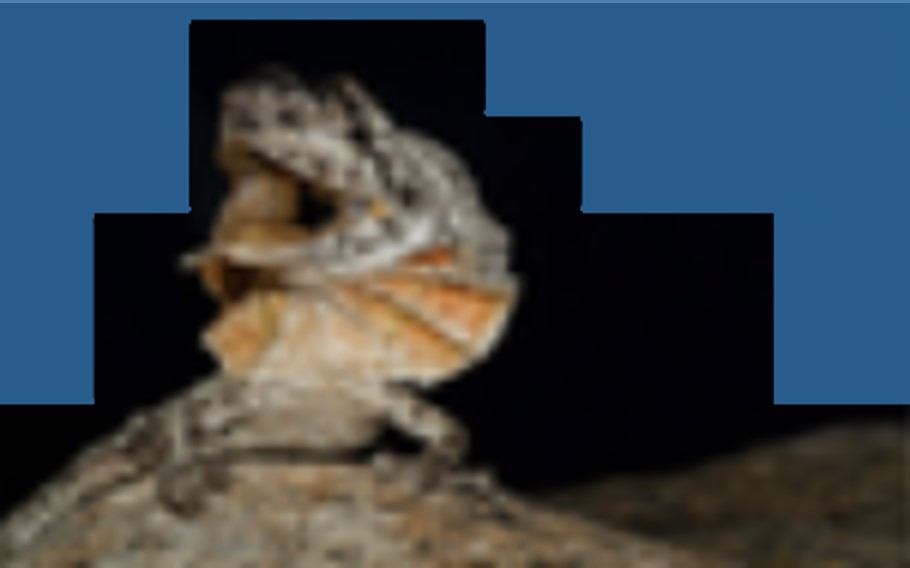}
        \caption{}
        \label{subfig:imagenet-vgg}
      \end{subfigure}
  % \hfill
  % \begin{subfigure}{0.53\linewidth}
  %   \includegraphics[width=\linewidth]{images/ComparisonOther.jpg}
  %   \caption{}
  %   \label{subfig:imgclspercentages}
  % \end{subfigure}
    
  \caption{The proposed technique can ignore regions of irrelevant pixels (marked in blue in the thresholded saliency maps) from the original images. The resulting Area of Interest (AoI) focuses on parts of the image that the human eye would. \textbf{(a)} Original COCO image. \textbf{(b)} fFaster-RCNN. \textbf{(c)} fSSDLite. \textbf{(d)} Original ImageNet image. \textbf{(e)} fResNet-18. \textbf{(f)} fVGG-16.  By ignoring computation on the blue regions, fCNNs save up to \textbf{12\% energy on ImageNet and up to 22\% energy on COCO.}
}
  \label{fig:example-aois}
\end{figure}

The qualitative results are positive as well. In \autoref{fig:example-aois}, we show examples of the AoIs selected by our $\tau$-thresholds in the different CNNs on images from COCO and ImageNet. Often, the selected AoIs draw the CNN's focus to the same areas that human eyes would focus on, although sometimes, the pretrained CNN seems to focus on areas of the image that seem less relevant. As shown, the technique can identify multiple AoIs in the pictures. 

% \begin{figure}[t]
%     \centering
%     \includegraphics[width=\linewidth]{images/imgclspercentages.jpg}
%     \caption{ImageNet accuracy VS energy consumption of the fCNN, relative to the unmodifed CNN.
%     % Compared to the unmodified CNNs, selecting focused convolution AoI thresholds for improved latency, and thus reduced energy consumption, causes accuracy to begin degrading. However, it does not begin degrading immediately, allowing our technique to achieve about an \textbf{8\%-12\% energy consumption improvement without accuracy degradation (top right data points)}. 
%     % The proposed technique improves more rapidly than accuracy degrades: 
%     % Right: More Energy (worse), Top: More accuracy (better)
%     % \todo[inline]{YHL: This figure is difficult to read you mirror both axes. Can you draw energy-accuracy, not reduction?}
%     }
%     
% \end{figure}

 A notable regression is fSSDLite. The Torchvision pretrained SSDLite model is noted as more sensitive to perturbations in pixel values, so we suspect that the deletion of pixels marked irrelevant still negatively impacts the model.
 Additionally, SSDLite does not use the multi-scale Feature Pyramid Network from the Torchvision Faster-RCNN~\cite{fasterrcnn}, struggling more when objects are smaller.
 
We also note that as more aggressive $\tau$ thresholds are selected, the energy consumption of the models improves more quickly than the accuracy degrades; for a more extreme example, it is possible to achieve a 28\% energy consumption improvement on ConvNeXt-T with only a 15\% loss in accuracy (accuracy dropped by 0.003, from 0.821 to 0.818).

This technique allows accuracy and latency to be traded off as the deployment scenario requires.

\subsection{Comparison with Similarly Inspired Techniques}
While not an apples-to-apples comparison since our technique does not require the training that the other methods do, we provide a comparison with similarly inspired techniques (\autoref{sec:background}) and an INT8-quantized baseline for ResNet-18.

Our focused fResNet-18 is both faster and more accurate than the Early Exit neural network~\cite{branchynet}, is faster than the standard ResNet-18~\cite{resnet}, is more accurate than quantized ResNet-18~\cite{pytorchquant}, and is faster than the Spatially Adaptive Computation model (SAC)~\cite{sact}.
Although quantized ResNet-18 is shown to be roughly 15ms faster than our focused ResNet-18, this speedup requires a time-consuming, difficult-to-understand calibration process on the dataset. Meanwhile, the focused ResNet-18 achieves its speedup without any such training.

\begin{figure}[t]
    \centering
    \includegraphics[width=\linewidth]{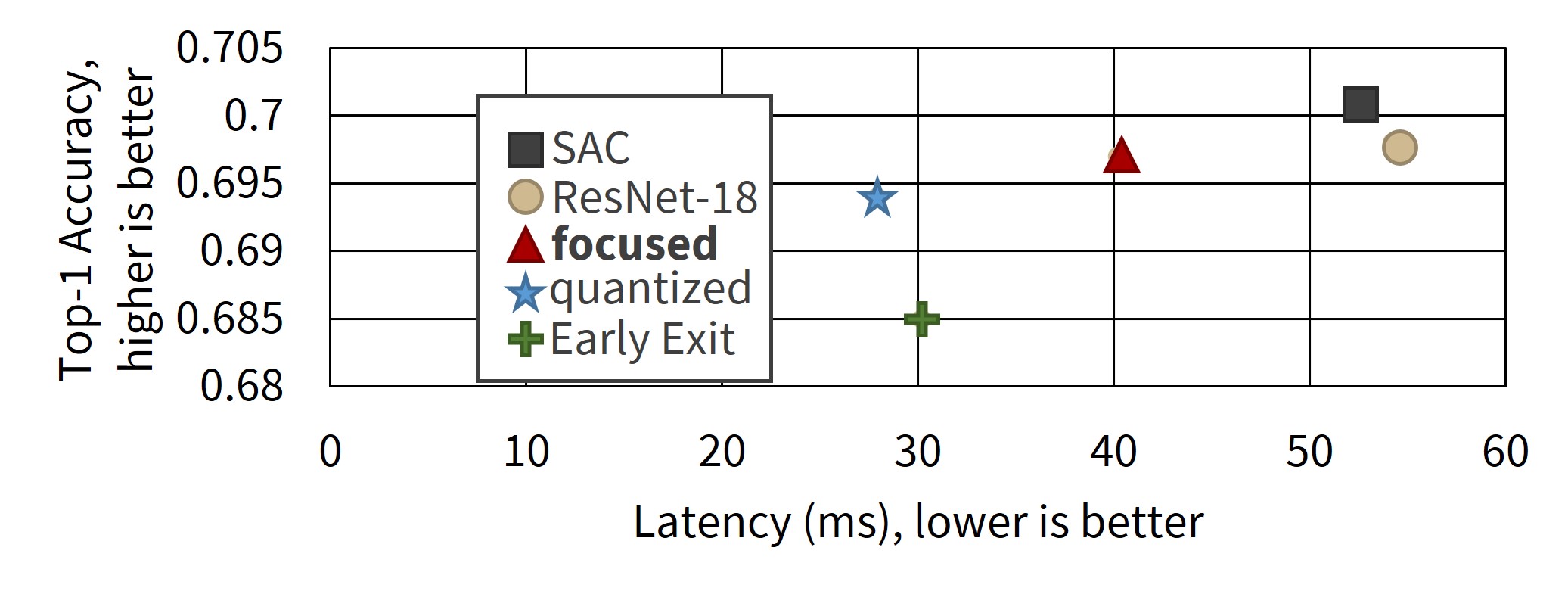}
    \caption{Our technique ``(focused)'' compared with similarly inspired techniques on our Intel CPU. SAC~\cite{sact} and Early Exit~\cite{branchynet} CNNs require training and a complete redesign of the CNN; \textbf{our technique not only either beats them or stays competitive, but it also requires zero training, keeping the pretrained CNN intact.} Static-quantized~\cite{pytorchquant} and unmodified ResNet-18 are shown to provide a baseline reference.}
    \label{fig:sota-comparison}
\end{figure}

In short, our technique either outperforms or stays competitive with similarly inspired techniques (\autoref{fig:sota-comparison}), all while being easy to implement because it requires no training.

\section{Conclusion}
\label{sec:conclusion}
This paper presents a novel technique for converting a pretrained computer vision model into a more energy-efficient model, with no additional training.
% We observe that highly accurate CNNs are typically accompanied by high energy consumption.
% We propose a technique to convert a CNN into a more energy-efficient model without requiring additional training.
We apply a threshold (determined using an accuracy-latency curve search method) to the features produced by the few early layers of the CNN to automatically generate an Area of Interest (AoI) for the given input image.
Pixels inside the AoI are relevant, the rest are irrelevant.
Irrelevant pixels are ignored, reducing computational cost and energy expenditure while improving inference latency.
The proposed technique uses a memory alignment method to ensure full utilization of parallel processing.
By keeping the weights and biases of the original pretrained model, a CNN pretrained on one dataset can still use our method for computation savings on a different dataset.
We achieve an average of 8\%-12\% energy savings with popular image classifiers (VGG, ConvNeXt) on ImageNet, and 15\%-22\% energy savings with popular object detectors (Faster-RCNN, RetinaNet) on COCO, with little to no loss in accuracy.
 Code is open-sourced on GitHub at https://github.com/PurdueCAM2Project/focused-convolutions.
% Since the accuracy degradation is low relative to the energy improvements, our future work should focus on recovering the lost accuracy.

\section*{Acknowledgements}

This research project is supported by funding from Cisco and National Science Foundation 2107230, 2107020, 2104709, 2104319, 2104377. Any opinions, findings, and conclusions expressed in this paper are those of the authors and do not necessarily reflect the sponsors' views.

{\footnotesize
\bibliographystyle{ieeetr}
\bibliography{manualbib}
}

\end{document}